\documentclass{article}

\usepackage{microtype}
\usepackage{graphicx}
\usepackage{subfigure}
\usepackage{booktabs} 

\usepackage{hyperref}

\usepackage[accepted]{icml2023}

\usepackage{amsmath}
\usepackage{amssymb}
\usepackage{mathtools}
\usepackage{amsthm}
\usepackage{inconsolata}
\usepackage{url}
\usepackage{multirow}
\usepackage{listings}
\usepackage{rotating}
\usepackage{vcell}
\usepackage{graphicx}
\usepackage{amsmath}
\usepackage{amsthm}
\usepackage{booktabs}
\usepackage{times}
\usepackage{soul}
\usepackage{url}
\usepackage{makecell}
\usepackage{color}
\usepackage{arydshln}
\usepackage{graphicx}
\usepackage{multirow}
\usepackage{array}
\usepackage{lipsum}
\usepackage{multicol}
\usepackage{color}
\usepackage[capitalize,noabbrev]{cleveref}

\theoremstyle{plain}

\theoremstyle{definition}

\theoremstyle{remark}

\usepackage[textsize=tiny]{todonotes}

\icmltitlerunning{Submission and Formatting Instructions for ICML 2023}

\begin{document}

\twocolumn[
\icmltitle{Data-Driven Approach for Formality-Sensitive Machine Translation: Language-Specific Handling and Synthetic Data Generation}




\begin{icmlauthorlist}
\icmlauthor{Seugnjun Lee}{yyy}
\icmlauthor{Hyeonseok Moon}{yyy}
\icmlauthor{Chanjun Park}{yyy,yyyc}
\icmlauthor{Heuiseok Lim}{yyy}
\end{icmlauthorlist}

\icmlaffiliation{yyy}{Department of Computer Science and Engineering, Korea University, Seoul 02841, Korea}
\icmlaffiliation{yyyc}{Upstage, Gyeonggi-do, Korea}

\icmlcorrespondingauthor{Heuiseok Lim}{limhseok@korea.ac.kr}

\icmlkeywords{Machine Learning, ICML}

\vskip 0.3in
]



\printAffiliationsAndNotice{}  

\begin{abstract}
In this paper, we introduce a data-driven approach for Formality-Sensitive Machine Translation (FSMT) that caters to the unique linguistic properties of four target languages. Our methodology centers on two core strategies: 1) language-specific data handling, and 2) synthetic data generation using large-scale language models and empirical prompt engineering. This approach demonstrates a considerable improvement over the baseline, highlighting the effectiveness of data-centric techniques. Our prompt engineering strategy further improves performance by producing superior synthetic translation examples.
\end{abstract}

\section{Introduction}
Neural machine translation (NMT) models, despite their impressive progress, often overlook the role of style and pragmatic aspects in translation, such as formality or politeness~\cite{britz2017massive,stahlberg2020neural}. This has given rise to the field of formality-sensitive machine translation (FSMT), which aims to control the level of formality in translated text across languages.

However, managing formality in MT is a challenging endeavor due to the lack of gold standard translations with different formality levels and the diverse formality markers across languages~\cite{nuadejde2022cocoa}. For example, in many Indo-European languages, personal pronouns and verb agreement denote formality. Meanwhile, in Korean, formality control is complex due to the common use of morphological markers to express polite, respectful, and humble speech, making it an intriguing test case for FSMT.

In this paper, we propose a data-centric approach to FSMT for the English-Korean (EN-KO) and English-Vietnamese (EN-VI) language pairs. Our approach comprises two primary strategies: 1) a language-specific data-driven technique, and 2) synthetic data generation using large-scale language models and prompt engineering.

\begin{table*}[]
\small
\centering
\renewcommand{\arraystretch}{1.0}
\resizebox{0.73\textwidth}{!}{%
\begin{tabular}{@{}llcccccccc@{}}
\toprule
 &  & \multicolumn{4}{c}{\textbf{EN-KO}} & \multicolumn{4}{c}{\textbf{EN-VI}} \\
 & \textbf{\textsc{Method}} & \multicolumn{1}{c}{\textsc{BLEU}} & \multicolumn{1}{c}{\textsc{COMET}} & \multicolumn{1}{c}{\textsc{\%M-Acc}} & \multicolumn{1}{c}{\textsc{\%C-F}} & \multicolumn{1}{c}{\textsc{BLEU}} & \multicolumn{1}{c}{\textsc{COMET}} & \multicolumn{1}{c}{\textsc{\%M-Acc}} & \multicolumn{1}{c}{\textsc{\%C-F}} \\ \midrule
\multirow{4}{*}{\textit{\textbf{\rotatebox[origin=c]{90}{\textit{Formal}}}}} & Official Baseline & 4.91 & 0.211 & 78.3 & 98.6 & 26.71 & 0.363 & 96.0 & 99.7 \\
 & ChatGPT & 5.65 & 0.524 & 83.3 & \textbf{100.0} & 27.07 & 0.510 & \textbf{100.0} & 98.0 \\
 & Ours & \textbf{26.60} & \textbf{0.727} & \textbf{87.0} & \textbf{100.0} & \textbf{47.00} & \textbf{0.669} & 99.4 & \textbf{100.0} \\
 & Ours + Augmentation & 17.09 & 0.667 & 79.4 & 99.5 & 41.57 & 0.653 & 99.4 & 99.7 \\ \midrule
\multirow{4}{*}{\textit{\textbf{\rotatebox[origin=c]{90}{\textit{Informal}}}}} & Official Baseline & 4.85 & 0.170 & 97.6 & 99.5 & 25.28 & 0.345 & 96.0 & 98.2 \\
 & ChatGPT & 5.60 & 0.564 & \textbf{100.0} & \textbf{100.0} & 25.83 & 0.482 & \textbf{100.0} & \textbf{100.0} \\
 & Ours & \textbf{27.10} & \textbf{0.715} & 98.0 & 95.0 & \textbf{45.60} & \textbf{0.637} & 98.8 & \textbf{100.0} \\
 & Ours + Augmentation & 20.35 & 0.621 & 98.5 & 98.8 & 40.46 & 0.484 & 98.7 & \textbf{100.0} \\ \bottomrule
\end{tabular}%
}
\caption{Results on the test set of Formality Dataset for formal and informal supervised settings.}
\label{tab:result_supervised}
\end{table*}

\begin{table*}[]
\small
\centering
\renewcommand{\arraystretch}{1.2} 
\resizebox{0.73\textwidth}{!}{%
\begin{tabular}{llcccccccc}
\toprule
 &  & \multicolumn{4}{c}{\textbf{EN-PT}} & \multicolumn{4}{c}{\textbf{EN-RU}} \\
 & \textbf{\textsc{Method}} & \multicolumn{1}{c}{\textsc{BLEU}} & \multicolumn{1}{c}{\textsc{COMET}} & \multicolumn{1}{c}{\textsc{\%M-Acc}} & \multicolumn{1}{c}{\textsc{\%C-F}} & \multicolumn{1}{c}{\textsc{BLEU}} & \multicolumn{1}{c}{\textsc{COMET}} & \multicolumn{1}{c}{\textsc{\%M-Acc}} & \multicolumn{1}{c}{\textsc{\%C-F}} \\ \hline
\multirow{3}{*}{\textit{\textbf{\rotatebox[origin=c]{90}{\textit{Formal}}}}} & Official Baseline & 27.29 & 0.448 & 96.3 & 97.7 & 21.96 & 0.349 & 96.2 & 92.0 \\
 & ChatGPT & \textbf{31.25} & \textbf{0.655} & 92.0 & 96.0 & \textbf{31.25} & 0.655 & 92.0 & 96.0 \\
 & Ours & 31.00 & 0.525 & \textbf{100.0} & \textbf{100.0} & 25.80 & 0.445 & \textbf{100.0} & \textbf{100.0} \\ \hline
\multirow{3}{*}{\textit{\textbf{\rotatebox[origin=c]{90}{\textit{Informal}}}}} & Official Baseline & \textbf{30.93} & 0.416 & \textbf{93.2} & \textbf{90.8} & 21.63 & 0.348 & 84.1 & 85.2 \\
 & ChatGPT & 27.38 & \textbf{0.512} & 48.4 & 46.0 & \textbf{31.25} & \textbf{0.655} & 92.0 & \textbf{100.0} \\
 & Ours & 19.90 & 0.249 & 68.0 & 90.0 & 26.30 & 0.418 & \textbf{100.0} & \textbf{100.0} \\ \bottomrule
\end{tabular}%
}
\caption{Results on the test set of Formality Dataset for formal and informal zero-shot settings.}
\label{tab:result_zeroshot}
\end{table*}

\section{Proposed Method}

\subsection{Language Specialized Data-Centric Approach}
We employ a language-specialized, data-centric approach that merges transfer learning techniques~\cite{zoph2016transfer} with language-specific subword methods, resulting in improved translation performance~\cite{zoph2016transfer, bojanowski2017enriching,park2020empirical,park2021should}. The pre-trained model (PLM) is fine-tuned on the supervised train set for each language pair.

For both EN-KO and EN-VI translations, we adopt a data-centric approach emphasizing pre-training and fine-tuning on high-quality, language-specific datasets. For EN-KO, we use a Transformer model and a morpheme-aware subword tokenization method~\cite{park2020empirical}, enhancing performance by addressing linguistic peculiarities of Korean. Similarly, for EN-VI, we utilize the specialized EnViT5~\cite{ngo2022mtet} model, with training conducted on the expanded CC100~\cite{wenzek-etal-2020-ccnet}, MTet~\cite{ngo2022mtet}, and PhoMT~\cite{doan-etal-2021-phomt} datasets, improving translation in low-resource settings and underrepresented domains.

\subsection{Synthetic Data Generation and Data-Centric Approach}
To enhance translation quality, especially in low-resource settings, we utilize a data-centric approach by generating synthetic examples using ChatGPT with the GPT-4 engine~\cite{openai2023gpt4}. Our synthetic data are created through a conditioned translation generation task and refined using a formality classifier~\cite{rippeth2022controlling}, thus ensuring accurate formality control.

\paragraph{Supervised Setting}
We leverage a prompt-based method, incorporating $n$ randomly selected shots from the English training set of various language pairs for context. These prompts guide ChatGPT to produce translations in either informal or formal target language. The translated examples are filtered using a formality classifier, and those meeting the formality criteria are integrated into the training sets for EN-KO and EN-VI fine-tuning. This data augmentation strategy and its impact are further evaluated through comparative experiments.

\paragraph{Zero-shot Setting}
In the zero-shot scenarios (EN-PT and EN-RU), we generate synthetic examples using the \citet{Gopalakrishnan2019}. As in the supervised setting, prompts guide the model to produce translations in either formal or informal target language. The examples are filtered for accurate formality before being used to fine-tune the pre-trained multilingual translation model. This approach maximizes the model's generalization ability across languages and formality levels, demonstrating the utility of synthetic data in expanding pre-trained language models' capabilities.

\section{Experiments}
\subsection{Experimental Settings}
We conducted experiments using the Formality dataset~\cite{nuadejde2022cocoa} for the language pairs EN-\{KO, VI\} in the supervised setting and EN-\{PT, RU\} in the zero-shot setting. Prompt engineering was applied for EN-\{KO, VI\}, and synthetic examples were generated for fine-tuning on EN-\{PT, RU\}. Training details varied for each language pair, with EN-KO utilizing morpheme-aware tokenization and pre-training with a Transformer model. EN-\{VI, PT, RU\} pairs were fine-tuned using mBART-50~\cite{liu2020multilingual} model.

\subsection{Experimental Results}
Our data-centric approach yielded promising results, as evidenced in Table~\ref{tab:result_supervised} and Table~\ref{tab:result_zeroshot} for supervised and zero-shot settings, respectively. Our model, trained on the Formality Dataset, demonstrated near-perfect formality control, with high translation accuracy for most tasks, especially in the EN-KO and EN-VI language pairs. However, data augmentation with ChatGPT sometimes led to subpar performance, hinting at the requirement for more elaborate prompts considering formality control. Notably, the zero-shot EN-PT task results were significantly low, suggesting a need for specialized techniques for formality control per language pair and revealing a potential training data bias in ChatGPT.

\section{Conclusion}
We propose a data-centric approach for FSMT, incorporating language-specific techniques and synthetic data generation. Our approach achieves superior performance in EN-KO and EN-VI translations, delivering high-quality formality translations. While EN-PT informal exhibits lower performance, other pairs surpass the baseline, showcasing the translation capabilities of ChatGPT. For future work, we suggest exploring larger translation models, analyzing shot examples in more depth, employing linguistic-based data augmentation, and further investigating zero-shot transfer to enhance FSMT performance.

\clearpage
\newpage
\section*{Acknowledgements}
This work was supported by Institute for Information \& communications Technology Planning \& Evaluation(IITP) grant funded by the Korea government(MSIT) (No. 2022-0-00369, (Part 4) Development of AI Technology to support Expert Decision-making that can Explain the Reasons/Grounds for Judgment Results based on Expert Knowledge) and supported by the National Research Foundation of Korea(NRF) grant funded by the Korea government(MSIT)(No. 2022R1A5A7026673). The authors would like to acknowledge that this paper is an abstract version of the paper titled "Improving Formality-Sensitive Machine Translation using Data-Centric Approaches and Prompt Engineering," which was submitted to the Formality Control for Spoken Language Translation Shared Task at the International Workshop on Spoken Language Translation (IWSLT) 2023.

\nocite{langley00}

\bibliography{example_paper}
\bibliographystyle{icml2023}

\newpage
\appendix
\onecolumn

\end{document}